\documentclass[11pt]{article}
\usepackage{acl2014}
\usepackage{times}
\usepackage{url}
\usepackage[none]{hyphenat}
\usepackage{latexsym}
\usepackage{amsmath}
\usepackage[]{natbib}
\usepackage{float} 
\usepackage{bm}
\usepackage{graphicx}
\usepackage{caption}

\title{Sample-efficient Deep Reinforcement Learning for Dialog Control}
\author{Kavosh Asadi \\
Brown University \\
  {\tt kavosh@brown.edu} \\\And
 Jason D. Williams \\
Microsoft Research \\
  {\tt jason.williams@microsoft.com} \\}
\date{}

\begin{document}
\maketitle
\begin{abstract}
Representing a dialog policy as a recurrent neural network (RNN) is attractive because it handles partial observability, infers a latent representation of state, and can be optimized with supervised learning (SL) or reinforcement learning (RL).  For RL, a policy gradient approach is natural, but is sample inefficient.  In this paper, we present 3 methods for reducing the number of dialogs required to optimize an RNN-based dialog policy with RL.  The key idea is to maintain a second RNN which predicts the value of the current policy, and to apply experience replay to both networks.  On two tasks, these methods reduce the number of dialogs/episodes required by about a third, vs. standard policy gradient methods.
\end{abstract}

\section{Introduction}
\label{sec:intro}

We study the problem of using reinforcement learning (RL) to optimize a controller represented as a recurrent neural network (RNN).  RNNs are attractive because they accumulate sequential observations into a latent representation of state, and thus naturally handle partially observable environments, such as dialog, and also robot navigation, autonomous vehicle control, and others.  


Among the many methods for RL optimization \citep{sutton1998reinforcement}, we adopt the \emph{policy gradient} approach \citep{williams1992}.  Policy gradient approaches are a natural fit for recurrent neural networks because both make updates via stochastic gradient descent.  They also have strong convergence characteristics compared to value-function methods such as Q-learning, which can diverge when using function approximation \citep{precup2001off}.  Finally, the form of the policy makes it straightforward to also train the model from expert trajectories (ie, training dialogs), which are often available in real-world settings.

Despite these advantages, in practice policy gradient methods are often sample inefficient, which is limiting in real-world settings where explorational interactions -- ie, conducting dialogs -- can be expensive.

The contribution of this paper is to present a family of new methods for increasing the sample efficiency of policy gradient methods, where the policy is represented as a recurrent neural network (RNN).  Specifically, we make two changes to the standard policy gradient approach.  First, we estimate a \emph{second} RNN which predicts the expected future reward the policy will attain in the current state; during updates, the value network reduces the error (variance) in the gradient step, at the expense of additional computation for maintaining the value network.  Second, we add experience replay to both networks, allowing more gradient steps to be taken per dialog.   


This paper is organized as follows.  The next section reviews the policy gradient approach, Section \ref{sec:methods} presents our methods, Sections \ref{sec:prob1} and \ref{sec:prob2} present results on two tasks, Section \ref{sec:relatedwork} covers related work, and Section \ref{sec:concl} briefly concludes.

\section{Preliminaries}
\label{sec:prelim}

In a reinforcement learning problem, an agent interacts with a stateful environment to maximize a numeric reward signal.  Concretely, at a timestep $t$, the agent takes an action $a_{t}$, is awarded a real-valued reward $r_{t+1}$, and receives an observation vector $\mathbf{o}_{t+1}$.  The goal of the agent is to choose actions to maximize the discounted sum of future rewards, called the return, $G_t$. In an episodic problem, the return at a timestep $t$ is:
\begin{equation}
G_t = \sum_{i=1}^{T-t}\gamma^{i-1}  r_{t+i} \ ,
\end{equation}where $T$ is the terminal timestep, and $\gamma$ is a discount factor $0 \le \gamma \le 1$.

In this paper, we consider policies represented as a recurrent neural network (RNN).  Internally the RNN maintains a vector representing a latent state $\mathbf{s}$, and the latent state begins in a fixed state $\mathbf{s}_0$.  At each timestep $t = 1,2,\ldots$, an RNN takes as input an observation vector $\mathbf{o}_t$, updates its internal state according to a differentiable function $F(\mathbf{s}_{t-1},\mathbf{o}_t|\bm{\theta}_f) = \mathbf{s}_t$, and outputs a distribution over actions $\mathbf{a}_t$ according to a differentiable function $G(\mathbf{s}_t|\bm{\theta}_g) = \mathbf{a}_t$, where $\bm{\theta} = (\bm{\theta}_f,\bm{\theta}_g)$ parameterize the functions.  $F$ and $G$ can be chosen to implement long short-term memory (LSTM) \citep{lstm}, gated recurrent unit \citep{cho2014learning}, or other recurrent (or non-recurrent) models. $\pi^{\theta}(a_t|\mathbf{h}_t)$ denotes the output of the RNN at timestep $t$.

Past work has established a principled method for updating the parameters $\bm{\theta}$ of the policy $\pi^{\bm\theta}$ via RL \citep{williams1992,sutton2000,peters2006policy} via stochastic gradient descent:
\begin{equation}
\bigtriangleup \bm{\theta}_{d} = \sum_{t=0}^{T-1}\bigtriangledown_{\bm{\theta}} \log \pi^{\bm{\theta}}(a_t|\mathbf{h}_t) G_t\ .
\label{eq:REINFORCE_update_without_baseline}
\end{equation}
While this update is unbiased, in practice it has high variance and is slow to converge.  \cite{williams1992} and \cite{sutton2000} showed that this update can be re-written as
\begin{equation}
\bigtriangleup \bm{\theta}_{d} = \sum_{t=0}^{T-1}\bigtriangledown_{\bm{\theta}} \log \pi^{\theta}(a_t|\mathbf{h}_t) ( G_t - b),
\label{eq:REINFORCE_update_with_baseline}
\end{equation}
where $b$ is a baseline, which can be an arbitrary function of states visited in dialog $d$.  Note that this update assumes that actions are drawn from the same policy parameterized by $\theta$ -- ie, this is an \emph{on-policy} update.

Throughout the paper, we also use importance sampling ratios that enable us to perform off-policy updates. Assume that some behavior policy $\mu$ is used to generate dialogs, and may in general be different from the target policy $\pi$ we wish to optimize.  At timestep $t$, we define the importance sampling ratio as
\begin{equation}
\rho_t=\frac{\pi(a_t|\textbf{h}_t)}{\mu(a_t|\textbf{h}_t)}.  \label{eq:importance_sampling}
\end{equation}
\section{Methods}
\label{sec:methods}
\subsection{Benchmarks}
Before introducing our methods, we first describe our two benchmarks.  The first uses (\ref{eq:REINFORCE_update_without_baseline}) directly.  The second uses (\ref{eq:REINFORCE_update_with_baseline}), with $b$ computed as an estimate of the average return of $\pi$:
\begin{equation}
b=\frac{\sum_{d \in \mathcal{D}} w(d)G^{d}_0}{\sum_{d \in \mathcal{D}}w(d)}
\end{equation}
where $\mathcal{D}$ is a window of most recent episodes (dialogs), and the weight of each dialog $w(d)$ is $w(d)=\Pi_{t=0}^{T(d)-1}\rho_{t}$. To compute ratios using (\ref{eq:importance_sampling}), the policy that generated the data is $\mu$, and the current (ie, target) policy is $\pi$.

\subsection{Method 1: State value function as baseline}
Our first method modifies parameter update (\ref{eq:REINFORCE_update_with_baseline}) by using a per-timestep baseline:
\begin{equation}
\bigtriangleup \bm{\theta}_{d}^{\textrm{on}} = \sum_{t=0}^{T-1}\bigtriangledown_{\bm{\theta}} \log \pi^{\bm\theta}(a_t|\mathbf{h}_t) \big(G_t - \hat V(\textbf{h}_t,\mathbf w)\big), 
\label{eq:REINFORCE_update_with_baseline_value}
\end{equation}
where $\hat V(\textbf{h}_t,\mathbf w)$ is an estimate of the value of $\pi^{\bm\theta}$ starting from $\mathbf{h}_t$, and $\mathbf w$ parameterizes $\hat V$.  This method allows a gradient step to be taken in light of the value of the current state.  We implement $\hat V(\textbf{h}_t,\mathbf w)$ as a \emph{second} RNN, and update its parameters at the end of each dialog using supervised learning, as
\begin{equation}
\bigtriangleup \mathbf{w}^{\textrm{on}}_{d}=\sum_{t=0}^{T-1}\big( G_t-\hat V(\mathbf{h_t,w})\big)\bigtriangledown_{\mathbf{w}}\hat V(\mathbf{h_t,w})
\label{eq:value_update_on_policy}
\end{equation}
Note that this update is also on-policy since the policy generating the episode is the same as the policy for which we want to estimate the value. 

\subsection{Method 2: Experience replay for value network}
Method 2 increases learning speed further by reusing past dialogs to better estimate $\hat V(\mathbf{h_t,w})$.  Since the policy changes after each dialog, past dialogs are \emph{off-policy} with respect to $\hat V(\mathbf{h_t,w})$, so a correction to (\ref{eq:value_update_on_policy}) is needed.  \cite{precup2001off} showed that the following off-policy update is equal to the on-policy update, in expectation:
\begin{eqnarray*}
\bigtriangleup \mathbf{w}^{\textrm{off}}_{d} = &&\sum_{t=0}^{T-1}\underset{i=0}{\overset{t-1}{\Pi}}\rho_i\\
   &&\big( \underset{j=0}{\overset{T-1}{\Pi}}\rho_j G_{t}-\hat V(\mathbf{h_t,w})\big)\bigtriangledown_{\mathbf{w}}\hat V(\mathbf{h_t,w}).
\label{eq:off_policy_value}
\end{eqnarray*}
Our second method takes a step with $\bigtriangleup \mathbf{w}^{\textrm{on}}_{d}$ with $d$ as the last dialog, and one or more steps with $\bigtriangleup \mathbf{w}^{\textrm{off}}_{d}$ where $d$ is sampled from recent dialogs.

\subsection{Method 3: Experience replay for policy network}

Our third method improves over the second method by applying experience replay to the policy network. Specifically, \citet{degris2012off} shows that samples of the following expectation, which is under behavior policy $\mu$, can be used to estimate the gradient of the policy network representing the policy $\pi^\theta$:
\begin{equation}
\mathbf{E}[\rho_t \bigtriangledown_{\bm{\theta}} \log \pi^{\theta}(a_t|\mathbf{h}_t)Q^{\pi}(\mathbf{h}_t,a_t)|\mu]
\end{equation}

We do not have access to $Q^{\pi}$, but since $Q^{\pi}(\mathbf{h}_t,a_t)=r_t+\gamma V^{\pi}(\mathbf{h}_{t+1})$, we can state the following off-policy update for the policy network:
\begin{eqnarray*}
\bigtriangleup \bm{\theta}^{\textrm{off}}_{d} =&& \sum_{t=0}^{T-1}\rho_t \bigtriangledown_{\bm{\theta}} \log \pi^{\theta}(a_t|\mathbf{h}_t)\\
&&\big(r_t+\gamma \hat V(\mathbf{h_{t+1},w})-\hat V(\mathbf{h_{t},w})\big)
\end{eqnarray*}
Method 3 first applies Method 2, then updates the policy network by taking one step with $\bigtriangleup \bm{\theta}^{\textrm{on}}_{d}$, followed by one or more steps with $\bigtriangleup \bm{\theta}^{\textrm{off}}_{d}$ using samples from recent dialogs.

\section{Problem 1: dialog system}
To test our approach, we created a dialog system for initiating phone calls to a contact in an address book, taken from the Microsoft internal employee directory.  Full details are given in \cite{WilliamsArchive16}; briefly, in the dialog system, there are three entity types -- \texttt{name}, \texttt{phonetype}, and \texttt{yesno}.  A contact's name may have synonyms (``Michael''/``'Mike'') and a contact may have more than one phone types (eg ``work'', ``mobile'') which in turn have synonyms (eg ``cell'' for ``mobile'').  

To run large numbers of RL experiments, we then created a stateful goal-directed simulated user, where the goal was sometimes not covered by the dialog system, and where the behavior was stochastic -- for example, the simulated user usually answers questions, but can instead ignore the system, provide additional information, or give up.  The user simulation was parameterized with around 10 probabilities.  

We defined the reward as $+1$ for successfully completing the task, and $0$ otherwise.  $\gamma=0.95$ was used to incentivize the system to complete dialogs faster rather than slower.  For the policy network, we defined $F$ and $G$ to implement an LSTM with 32 hidden units, with a dense output layer with a softmax activation.  The value network was identical in structure except it had a single output with a linear activation.  We used a batch size of 1, so we update both networks after completion of a single dialog. We used Adadelta with stepsize $\alpha=1.0,\ \rho=0.95,$ and $\epsilon=10^{-6}$. Dialogs took between 3 and 10 timesteps.  Every 10 dialogs, the policy was frozen and run for 1000 dialogs to measure average task completion.  
\label{sec:prob1}
\begin{figure}
\centering
\includegraphics[width=.9\columnwidth]{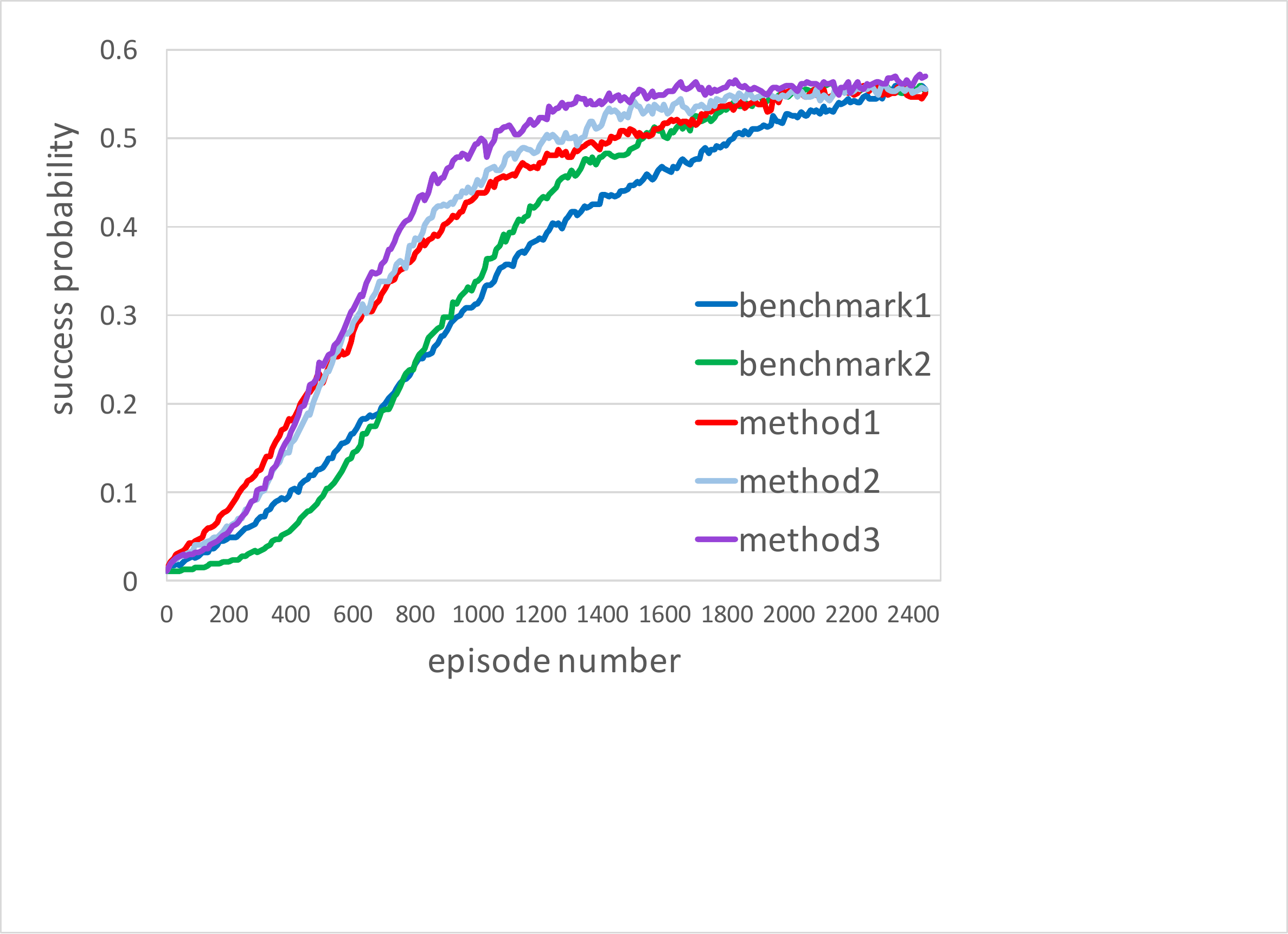}
\caption{Number of dialogs vs. average task success over 200 runs for the dialog task.}
\label{fig:dialog_mean}
\end{figure}

\begin{figure}
\centering
\includegraphics[width=.9\columnwidth]{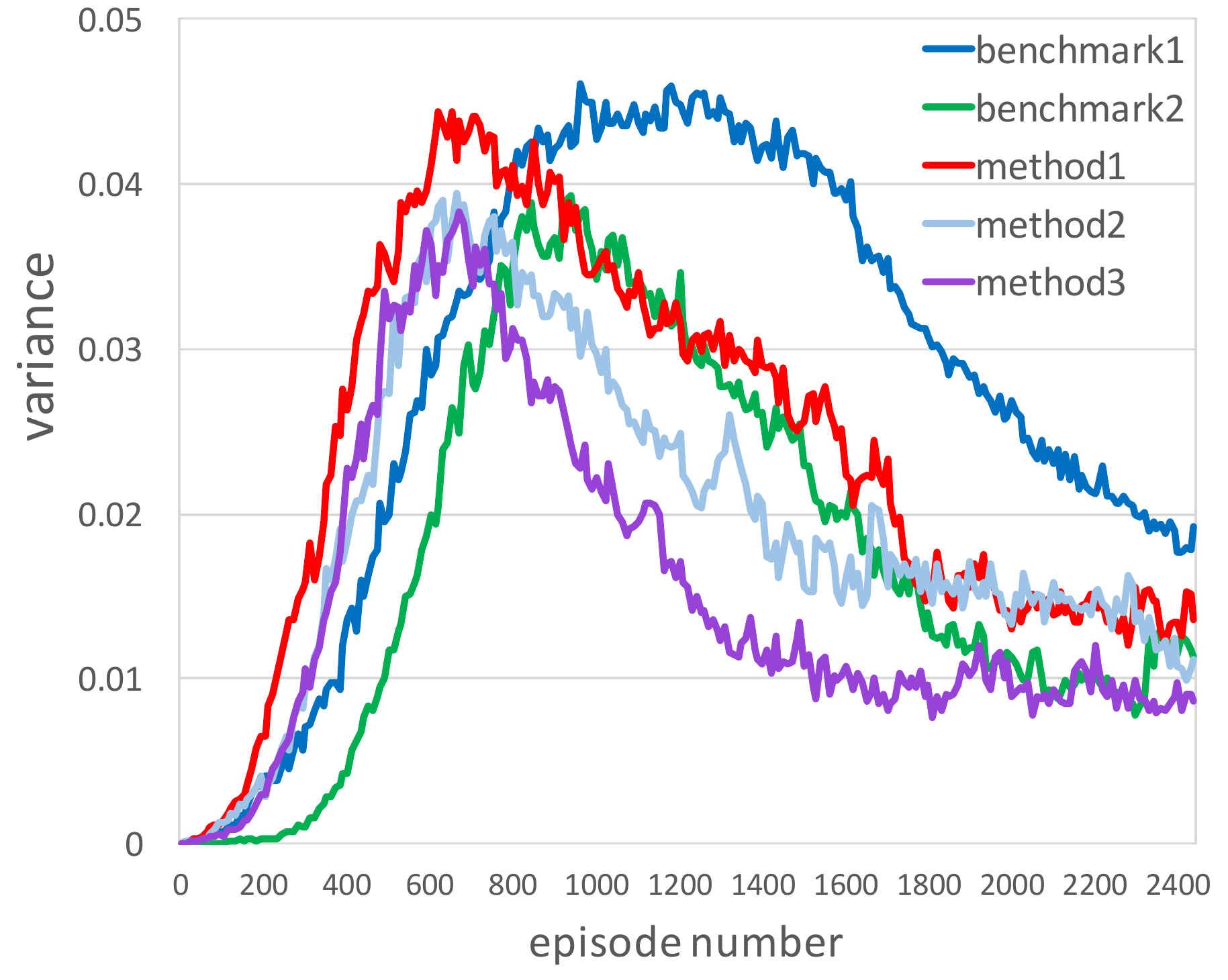}
\caption{Number of dialogs vs. variance in task success among 200 runs.}
\label{fig:dialog_variance}
\end{figure}


Figures \ref{fig:dialog_mean} and \ref{fig:dialog_variance} show mean and variance for task completion over 200 independent runs.  Compared to the benchmarks, our methods require about one third fewer dialogs to attain asymptotic performance, and have lower variance. 

\section{Problem 2: lunar lander}
\label{sec:prob2}
To test generality and provide for reproducibility, we sought to evaluate on a publicly available dialog task.  However, to our knowledge none exists, and so we instead applied our method to a public but non-dialog RL task, called ``Lunar Lander'', from the OpenAI gym.\footnote{\url{https://gym.openai.com/envs/LunarLander-v2}}  This domain has a continuous (fully-observable) state space in 8 dimensions (eg, x-y coordinates, velocities, leg-touchdown sensors, etc.), and 4 discrete actions (controlling 3 engines).  The reward is +100 for a safe landing in the designated area, and -100 for a crash. Using the engines will also result in a negative cost as explained in the link below. Episodes finish when the spacecraft crashes or lands.  We used $\gamma=0.99$.

Since this domain is fully observable, we chose definitions of $F$ and $G$ in the policy network corresponding to a fully connected neural networks with 2 hidden layers, followed by a softmax normalization.  We further chose RELU activations and 16 hidden units, based on limited initial experimentation.  The value network has the same architecture except for the output layer that has a single node with linear activation.  We used a batch episode size of 10, as we found that with a batch of 1 divergence appears frequently.  We used a stepsize of $0.005$ and used Adam algorithm \citep{kingma2014adam} with its default parameters.  Methods 2 and 3 performs 5 off-policy updates per each on-policy update for the value network.  Method 3 performs 3 off-policy updates per each on-policy update for the policy network. 

\begin{figure}
\centering
\includegraphics[width=.9\columnwidth]{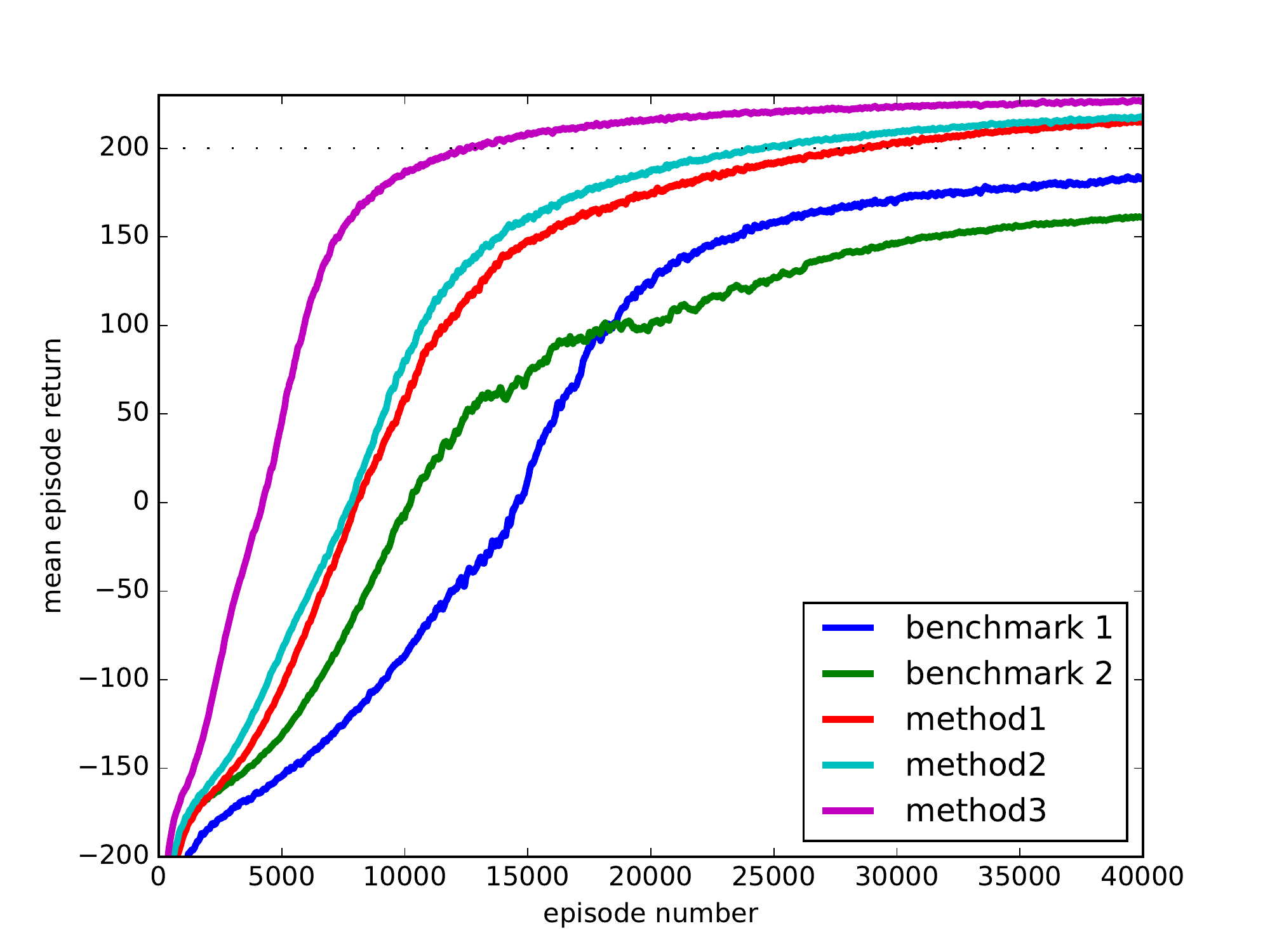}
\caption{Number of epsidoes vs. average return over 200 runs for the lunar lander task.}
\label{fig:lunar_lander}
\end{figure}

Results are in Figure \ref{fig:lunar_lander}, and show a similar increase in sample efficiency as in the dialog task.

\section{Related work} 
\label{sec:relatedwork}

Since neural networks naturally lend themselves to policy gradient-style updates, much past work has adopted this broad approach.  However, most work has studied the fully observable case, whereas we study the partially observable case.  For example, AlphaGo \citep{alphago} applies policy gradients (among other methods), but Go is fully observable via the state of the board.  

Several papers that study fully-observable RL are related to our work in other ways.  \cite{degris2012off} investigates off-policy policy gradient updates, but is limited to linear models.  Our use of experience replay is also off-policy optimization, but we apply (recurrent) neural networks.  Like our work, \citet{fatemi2016} also estimates a value network, uses experience replay to optimize that value network, and evaluates on a conversational system task.  However, unlike our work, they do not use experience replay in the policy network, their networks rely on an external state tracking process to render the state fully-observable, and they learn feed-forward networks rather than recurrent networks.

\citet{hausknecht2015} applies RNNs to partially observable RL problems, but adopts a Q-learning approach rather than a policy gradient approach.  Whereas policy gradient methods have strong convergence properties, Q-learning can diverge, and we observed this when we attempted to optimize Q represented as an LSTM on our dialog problem.  Also, a policy network can be pre-trained directly from (near-) expert trajectories using classical supervised learning, and in real-world applications these trajectories are often available \citep{WilliamsArchive16}.


\section{Conclusions}
\label{sec:concl}

We have introduced 3 methods for increasing sample efficiency in policy-gradient RL.  In a dialog task with partially observable state, our best method improved sample efficiency by about a third.  On a second fully-observable task, we observed a similar gain in sample efficiency, despite using a different network architecture, activation function, and optimizer.  This result shows that the method is robust to variation in task and network design, and thus it seems promising that it will generalize to other dialog domains as well.  In future work we will apply the method to a dialog system with real human users

\bibliographystyle{plainnat}
\bibliography{kavosh_rl_paper}
\end{document}